\def\year{2018}\relax
\documentclass[letterpaper]{article} 
\usepackage{aaai18}  
\usepackage{times}  
\usepackage{helvet}  
\usepackage{courier}  
\usepackage{url}  
\usepackage{graphicx}  
\usepackage{color}
\usepackage{amsmath}
\usepackage{amssymb}
\usepackage[normalem]{ulem}
\usepackage{algorithm}
\usepackage{algorithmicx}
\usepackage{algpseudocode}
\usepackage{mathtools}

\MakeRobust{\Call}

\title{Towards Online Learning from Corrective Demonstrations}
\author{Reymundo A. Gutierrez \\
            Department of Computer Science \\
            University of Texas at Austin \\
        \And Elaine Schaertl Short \\
            Department of Electrical and Computer Engineering \\
            University of Texas at Austin \\
        \AND Scott Niekum \\
            Department of Computer Science \\
            University of Texas at Austin \\
        \And Andrea L. Thomaz \\
            Department of Electrical and Computer Engineering \\
            University of Texas at Austin \\
       }

\begin{document}
%
\maketitle

\begin{abstract}

Robots operating in real-world human environments will likely encounter task execution failures. To address this, we would like to allow co-present humans to refine the robot's task model as errors are encountered. Existing approaches to task model modification require reasoning over the entire dataset and model, limiting the rate of corrective updates. We introduce the State-Indexed Task Updates (SITU) algorithm to efficiently incorporate corrective demonstrations into an existing task model by iteratively making local updates that only require reasoning over a small subset of the model. In future work, we will evaluate this approach with a user study.

\end{abstract}

\section{Introduction}


The dynamic and unstructured nature of real-world human environments precludes the modeling of all possible failure modes and their associated recovery policies.
To address this issue, we would like to allow co-present humans to correct the robot's behavior as execution failures are encountered. In particular, we are interested in developing techniques that enable naive users to provide corrections and improve the robot's performance over time.

Much prior work has separately investigated the learning and refinement of individual primitives \cite{akgun2012keyframe,sauser2012iterative,jain2013learning,Akgun2016,Bajcsy2018} and the sequencing of learned primitives \cite{kappler2015data}. Recent efforts have focused on learning task models by jointly reasoning over action primitives and their sequencing \cite{kroemer2015towards,niekum2015learning,Gutierrez2018IncrementalTM}. However, these methods present scalability problems as they require reasoning over the entire collected dataset or the entire current model which limits the rate of corrective updates.

In this work, we present State-Indexed Task Updates (SITU),
a new method that utilizes corrective demonstrations to efficiently update task models represented as a finite-state automaton (FSA) by making iterative local updates that only require reasoning over a small subset of the task model.
Given a segmented corrective demonstration, SITU first uses the world state to select the relevant subset of the task model.
Then, SITU determines if the demonstration segment is represented in the selected model subset, re-training and re-building model components as needed.
This process repeats until the final segment of the demonstration. 


\begin{figure}
	\centering
	\includegraphics[scale=0.45]{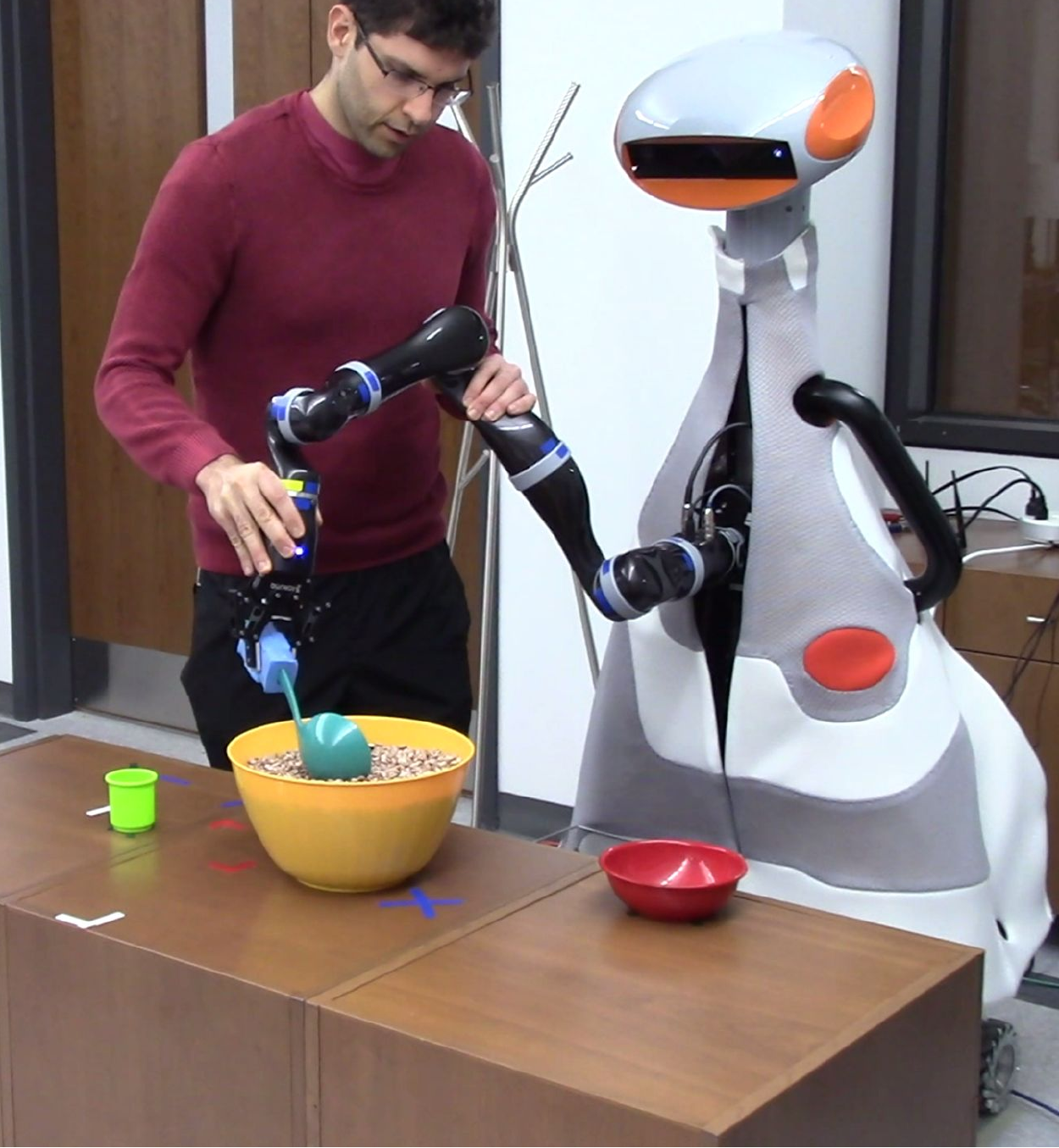}
	\caption{User teaching the robot by kinesthetic interaction}
	\label{fig:bowl_demo}
\end{figure}

\section{Related Work}

Much prior work has explored the topic of model refinement from corrections or subsequent demonstrations for individual primitives. Akgun et al. \cite{akgun2012keyframe} introduced a method for learning keyframe-based models of skills, and an interaction model to add/remove keyframes in the learned model in subsequent interactions. In work by Sauser et al. \cite{sauser2012iterative}, a teacher can provide tactile feedback to a robot during playback of a learned skill to adapt or correct object grasping behaviors. Jain et al. \cite{jain2013learning} introduced a method for iterative improvement of trajectories in order to incorporate user preferences, with new demonstrations providing information regarding task constraints. A similar approach was explored by Bajcsy et al. \cite{Bajcsy2018}, with user demonstrations being used to modify only one task constraint dimension at a time. Our approach seeks to simultaneously update many primitives and overall task structure from a set of extended demonstrations.

Various approaches have addressed the problem of updating both primitives and task structure. Kappler et al. \cite{kappler2015data} separate the two by learning a set of associative skill memories (ASMs) and manually-specifying a manipulation graph that sequences the ASMs. New ASMs are learned to incorporate recovery behaviors as needed and attached to the manipulation graph through user specification. Niekum et al. \cite{niekum2015learning} incrementally learn a finite state automaton (FSA) for a task by segmenting demonstrations with a beta-process auto-regressive hidden Markov model (BP-AR-HMM) and constructing a graph using the segment sequences. Classifiers are trained at branching points to determine which primitives to execute. Corrections can be provided after execution failure and incorporated into the graph by re-segmenting demonstrations and rebuilding the FSA. Gutierrez et al. \cite{Gutierrez2018IncrementalTM} provide a taxonomy of edit types on an FSA and learn corrections by instantiating models assuming a set of candidate corrections of each type. A state-based transitions autoregressive hidden Markov model (STARHMM) \cite{kroemer2015towards} is then trained on each of the candidate corrections using the provided corrective demonstrations, with the best model update selected according to a modified AIC score. Each of the above approaches present scalability problems particularly for the online case as they require either manual specification or evaluation on the entire dataset or model. Our approach leverages state information and execution history to limit the number of primitives to consider at any time step.

Mohseni-Kabir et al. \cite{mohseni2018simultaneous} simultaneously learn task hierarchy and primitives through interactive demonstrations. The demonstrator's narration is leveraged to identify boundaries between primitives and a set of heuristics are used to query the demonstrator regarding the grouping of primitives. This approach allows for online updates to the model. However, its applicability is limited as it requires the abstract structure of objects used (parts relations), full geometry to explore and learn constraints, as well as the re-targeting of motion captured human demonstrations onto the robot platform. Our approach gathers demonstrations directly on the robot and relies only on visual features to construct and update its models.

\section{Algorithm}

State-Indexed Task Updates (SITU) provides a method for incorporating corrective demonstrations, defined in this work as a partial demonstration given at or near execution failure. For any given task model, the corrective demonstration is a combination of modeled and unmodeled segments. Thus, incorporating the corrective demonstration into the existing task model requires determining which segments are unmodeled and adding/updating the associated components.

\subsection{Problem Formulation}


We model a task as a finite-state automaton (FSA), with nodes representing primitives and edges representing valid transitions between primitives. In this work, a node $z$ consists of a policy model ($\pi$) - describing when to take which action and an initiation classifier ($c$) - describing where a primitive can begin. More concretely, for each node

\begin{equation}
    z_i = \{ \pi_i(\boldsymbol{s}), c_i(\boldsymbol{s}) \}
\end{equation}

\noindent
where $i \in \{1, \ldots, \kappa\}$ and $\boldsymbol{s} \in \mathbb{R}^n$ is the observed state.

The edges of the graph are encoded in the following two functions:


\begin{itemize}
	\item $parent(z)$: returns the parents of node $z$
	\item $children(z)$: returns the children of node $z$
\end{itemize}

A full policy execution is computed by selecting a primitive and executing its policy. Primitive selection follows the structure of the graph such that if the current primitive is $z_i$, the most likely primitive from the set $Z = \{ z_j | j \in children(z_i) \}$ is selected as the next primitive. Using this task model, the problem of corrective updates is reduced to determining which policies and classifiers need to be updated and how the corrective demonstrations progress through the FSA (i.e. ordered set of traversed edges). 

The approach presented here is not limited to particular policy or classifier models. 
However, we require that the classifiers provide a measure of classification confidence. For the policies, we require that the following operator is available for the selected policy representation:

\begin{equation}
	\Call{Find-Policy}{d, \Pi} = \begin{cases}
	                                index(\pi) & \exists \pi \in \Pi, d \in \pi \\
	                                \varnothing & otherwise 
	                             \end{cases}
\end{equation}

\noindent
where $d$ is a demonstration segment and $\Pi$ is a set of policies. 
Given the policy set, \Call{Find-Policy}{} returns the index of the policy that the demo is an instance of, or returns empty if no such policy exists.
The implementation details of \Call{Find-Policy}{} are entirely dependent on the policy specification.

\subsection{State-Indexed Task Updates}
\label{sec:situ}



After segmenting the demonstration, \Call{SITU}{} iterates over the demonstration segments $D$ and their associated start states $S$ (Lines 3-5 of Algorithm \ref{alg:fsa_update}) and applies local updates to the FSA task model $T$ given the last executed primitive $a$, the current state $\boldsymbol{s}$, the current demonstration segment $d$, and the set of applicable primitives $Z$ (Lines 6-7 of Algorithm \ref{alg:fsa_update}). A primitive is considered applicable if its classifier is activated by the current state, which restricts the set of primitives used in subsequent steps.

\begin{algorithm}
    \caption{State-Indexed Task Updates} \label{alg:fsa_update}
    \begin{algorithmic}[1]
        \Procedure{SITU}{$z, S, D, T$}
            \State $a \gets z$
            \For{$i \gets 1$ \textbf{to} $|D|$}
                \State $\boldsymbol{s} \gets S[i]$
                \State $d \gets D[i]$
                \State $Z \gets \{ z_j \in T | c_j(\boldsymbol{s}) = 1 \}$
                \State $T, a \gets \Call{Local-Update}{a, \boldsymbol{s}, d, Z, T}$
            \EndFor
        \EndProcedure
    \end{algorithmic}
\end{algorithm}



The local update algorithm is outlined in Algorithm \ref{alg:local-update}. It makes use of a few helper functions that we will define here. \Call{Cluster}{} clusters the underlying demonstration segments in the nodes $Z$ and returns a new set of primitives retrained on the clustered data. \Call{Local-Reconnect}{} removes the old set of primitives $Z_{old}$ from the task model $T$ and connects the new primitives $Z_{new}$ according to the primitive traversal history of the underlying data. \Call{$\Pi$}{} returns the ordered set of policies for the set of nodes $Z$. \Call{S}{} returns the state samples used to train the classifiers of nodes $Z$. \Call{Fit-Policy}{} and \Call{Fit-Classifier}{} train a new policy and classifier models, respectively. \Call{Update-Policy}{} and \Call{Update-Classifier}{} update existing policy and classifier models given new data, respectively. Finally, \Call{Add-Node}{} and \Call{Add-Edge}{} add new nodes and edges to the task model, respectively.

\Call{Local-Update}{} takes as input the current primitive $a$, the current state $\boldsymbol{s}$, the current demonstration segment $d$, the set of applicable primitives $Z$, and the task model $T$. Using the \Call{Find-Policy}{} operator, \Call{Local-Update}{} determines if the current segment is an instance of one of the primitives in $Z$ (Line 3 of Algorithm \ref{alg:local-update}). If the current segment is not an instance of one of the primitives in $Z$, a new primitive is created and connected to the existing FSA (Lines~5-12~of~Algorithm \ref{alg:local-update}). If the current segment is an instance of one of the primitives in $Z$, that primitive is retrained with the new data the segment provides (Lines 14-20 in Algorithm~\ref{alg:local-update}).


\begin{algorithm}
    \caption{Local Update} \label{alg:local-update}
    \begin{algorithmic}[1]
        \Procedure{Local-Update}{$a, \boldsymbol{s}, d, Z, T$}
            \State $Z_{new} \gets \Call{Cluster}{Z}$
            \State $T \gets$ \Call{Local-Reconnect}{$T,Z,Z_{new}$}
            \State $i \gets$ \Call{Find-Policy}{$d$, \Call{$\Pi$}{$Z_{new}$}} \Comment{find d's policy}
            \If{$i = \varnothing$} \Comment{add new primitive}
                \State $p \gets \boldsymbol{s}$ \Comment{positive samples}
                \State $n \gets \Call{S}{children(a)}$ \Comment{negative samples}
                \State $\pi \gets$ \Call{Fit-Policy}{$d$}
                \State $c \gets$ \Call{Fit-Classifier}{$p,n$}
                \State $z \gets \{ \pi, c \}$ \Comment{create new primitive}
                \State \Call{Add-Node}{$T,z$} \Comment{add new node}
                \State \Call{Add-Edge}{$T,(a,z)$} \Comment{add new edge $a \rightarrow z$}
                \State \Return $T$, $z$
            \Else \Comment{update existing primitive}
                \State $z \gets Z_{new}[i]$ \Comment{get existing node}
                \State $p \gets \boldsymbol{s}$ \Comment{positive samples}
                \State $n \gets$ \Call{S}{$children(a) \cup$ \newline
                    \hspace*{5em} $\smashoperator[r]{\bigcup\limits_{r \in parent(z)}} children(r)$} \Comment{negative samples}
                \State \Call{Update-Policy}{$z,d$}
                \State \Call{Update-Classifier}{$z,p,n$}
                \State \Call{Add-Edge}{$T,(a,z)$} \Comment{add new edge $a \rightarrow z$}
                \State \Return $T$, $z$
            \EndIf
        \EndProcedure
    \end{algorithmic}
\end{algorithm}


The above description of Algorithm \ref{alg:local-update} left out the role of Lines 2 and 3. The FSA update algorithm as described so far can result in primitives that execute similar actions on similar regions of the state space. While this is not necessarily an impediment to successful execution, ideally we would like such redundant instances to be rolled into one primitive. Lines 2 and 3 of Algorithm \ref{alg:local-update} accomplish this by reconstructing the applicable primitive models with \Call{Cluster}{}, effectively merging similar policy models. Because this clustering step is limited to only a small subset of the nodes, the FSA can be quickly updated with \Call{Local-Reconnect}{}. This is equivalent to modifying the first primitive in the redundant set with the data of the newer primitives. 

\vspace{10pt}

In previous work, we defined a taxonomy of edit types that can be made to an FSA: node addition, edge addition, and node modification \cite{Gutierrez2018IncrementalTM}. It is important for an algorithm designed to to refine FSAs to perform these edit types as they represent the learning of new skills, learning of new skill transitions, and the refinement of previously learned skills. 
The above algorithms span the outlined edit types in the following manner. Node addition occurs during a local update when no sufficient policy model for the current demonstration segment is found (Lines 6-12 of Algorithm~\ref{alg:local-update}). Edge addition is performed in Lines 12 and 20 of Algorithm \ref{alg:local-update}. The edge addition is mediated through the indexing of primitives via their initiation classifiers (Line 6 of Algorithm \ref{alg:fsa_update}) and policy matching through the \Call{Find-Policy}{} function (Line 4 of Algorithm \ref{alg:local-update}). Finally, node modification occurs indirectly through the clustering (Line 2) and reconnecting (Line~3) steps of Algorithm \ref{alg:local-update}, as described in the previous paragraph.


\section{Keyframe Based Algorithms}

Prior work has shown that novice teachers can more successfully demonstrate tasks by providing keyframe demonstrations as opposed to trajectory demonstrations \cite{akgun2012keyframe}. Keyframes produce lower-noise demonstrations and achieve more consistent results. Motivated by this, we instantiate SITU for keyframe based policies.

\subsection{Segmentation}

A variety of algorithms are available to produce the demonstration segments required as input to SITU. While we focus on keyframe demonstrations and utilize keyframes as segmentation points in this work, the presented approach can be used with any segmentation algorithm. However, we note that the overall performance is dependent on the consistency of segmentation across demonstrations.



One approach for segmenting keyframe demonstrations is to produce a trajectory for each demonstration and use any of the several existing trajectory segmentation algorithms. Another approach, used in this work, is to use the changes in reference object as segmentation points. We assume in this work that the reference object is provided by the demonstrator. This can be obtained from demonstration narration (e.g.~``get the cup'') and has been accomplished in prior work \cite{mohseni2018simultaneous}. Thus, this approach produces segments that are a sequence of keyframes.

\subsection{Primitive Specification}


The primitives consist of a policy model and classifier model. For the classifier, we use a logistic regression model. Under this model, classification confidence is given in the form of a probability. The policy takes the form of a hidden Markov model (HMM) with multivariate Gaussian emissions over the end-effector pose relative to the reference object. The HMM is trained on a set of keyframe sequences, with each hidden state interpreted to correspond to an underlying `true' keyframe. The policy is executed by sampling a keyframe trajectory and planning between subsequent keyframes.

\subsection{Find Policy}


Algorithm \ref{alg:local-update} specifies separate \Call{Cluster}{} and \Call{Find-Policy}{} steps. For our keyframe instantiation of SITU, we combine them. In order to find the policy of the demonstration segment $d$ in the policy set $\Pi$, we first train an HMM over the trajectory that results from linearly interpolating between the keyframes of $d$. We then cluster this new HMM and the HMM policies in $\Pi$ using the average KL divergence between policy pairs as the distance function. Whichever cluster contains $d$'s HMM is its policy membership. Each cluster can then be used to train a new primitive, which are connected to the existing graph according to the primitive traversal history contained in the training data.








\section{Proposed Evaluation}

The described algorithms will be evaluated with a user study. Participants with no prior experience teaching a robot will be recruited from the local university campus. The participants will first familiarize themselves with keyframe teaching by performing a few simple exercises. After this practice session, they will provide demonstrations for the completion of a set of tasks. Examples include pouring from a pitcher and scooping contents from one bowl to another. The latter is shown in Figure~\ref{fig:bowl_demo}. The tasks will then be modified such that a correction is needed for successful execution (e.g. adding a lid removal step to the pouring task). The task re-specifications will be structured to span the set of edit types outlined in Section \ref{sec:situ}.

The performance will be evaluated by comparing the number and length of the demonstrations needed to successfully correct the task models against the number and length of demonstrations needed to construct a successful task model from scratch. The models constructed from all collected demonstrations (original+correction) will also be compared against the models that result from the separate corrective~steps.


\section{Conclusion}

Robots operating in unstructured environments will likely encounter task execution failures. To address this, we would like to allow co-present humans to correct the robot's behavior as errors are encountered. Existing approaches to task model modification require reasoning over the entire dataset and/or model, which limits the rate of corrective updates and presents interactive bottlenecks. We introduce the State-Indexed Task Updates (SITU) algorithm to efficiently incorporate corrective demonstrations into an existing task model by iteratively making local updates that only require reasoning over a small subset of the task model. In future work, we will evaluate this approach with a user study. 

\section*{Acknowledgments}
This work has taken place in the Socially Intelligent Machines (SIM) lab and the Personal Autonomous Robotics Lab (PeARL) at The University of Texas at Austin. This research is supported in part by the National Science Foundation (IIS-1638107, IIS-1724157) and the Office of Naval Research (\#N000141410003).

\bibliography{references}
\bibliographystyle{aaai}

\end{document}